\pdfoutput=1

\documentclass[11pt]{article}

\usepackage[review]{acl}

\usepackage{times}
\usepackage{latexsym}

\usepackage[T1]{fontenc}

\usepackage[utf8]{inputenc}
\usepackage{amsmath}
\usepackage{microtype}
\usepackage{multirow}
\usepackage{inconsolata}
\usepackage{hyperref}
\usepackage{pifont}
\title{Hierarchical Attention Graph for Scientific Document Summarization in Global and Local Level}

\author{Chenlong Zhao\textsuperscript{1,2} , Xiwen Zhou\textsuperscript{1,2} , Xiaopeng Xie\textsuperscript{1,2}, Yong Zhang\textsuperscript{1,2}\thanks{Corresponding author} \\
        \textsuperscript{1}School of Electronic Engineering,\\ Beijing University of Posts and Telecommunications, Beijing 100876, China\\ \textsuperscript{2}Beijing Key Laboratory of Work Safety Intelligent Monitoring,\\ Beijing University of Posts and Telecommunications, Beijing 100876, China\\ 
        \texttt{\{Chenlong\_000325,zhouxiwen,657314qq,yongzhang\}@bupt.edu.cn}}

\usepackage{graphicx}
\begin{document}
\maketitle
\begin{abstract}
Scientific document summarization has been a challenging task due to the long structure of the input text. The long input hinders the simultaneous effective modeling of both global high-order relations between sentences and local intra-sentence relations which is the most critical step in extractive summarization. However, existing methods mostly focus on one type of relation, neglecting the simultaneous effective modeling of both relations, which can lead to insufficient learning of semantic representations. In this paper, we propose HAESum, a novel approach utilizing graph neural networks to locally and globally model documents based on their hierarchical discourse structure. First, intra-sentence relations are learned using a local heterogeneous graph. Subsequently, a novel hypergraph self-attention layer is introduced to further enhance the characterization of high-order inter-sentence relations. We validate our approach on two benchmark datasets, and the experimental results demonstrate the effectiveness of HAESum and the importance of considering hierarchical structures in modeling long scientific documents\footnote{Our code will be available at \url{https://github.com/MoLICHENXI/HAESum}}.
\end{abstract}

\section{Introduction}

Extractive summarization aims to select a set of sentences from the input document that best represents the information of the whole document. With the advancement of pre-trained models and neural networks over the years, researchers have achieved promising results in news summarization \citep{liu2019text,zhong2020extractive}. However, when applying these methods to long scientific documents, they encounter challenges due to the relatively lengthy inputs. \begin{figure}[htbp]
\centering
    \includegraphics[width=80mm, keepaspectratio]{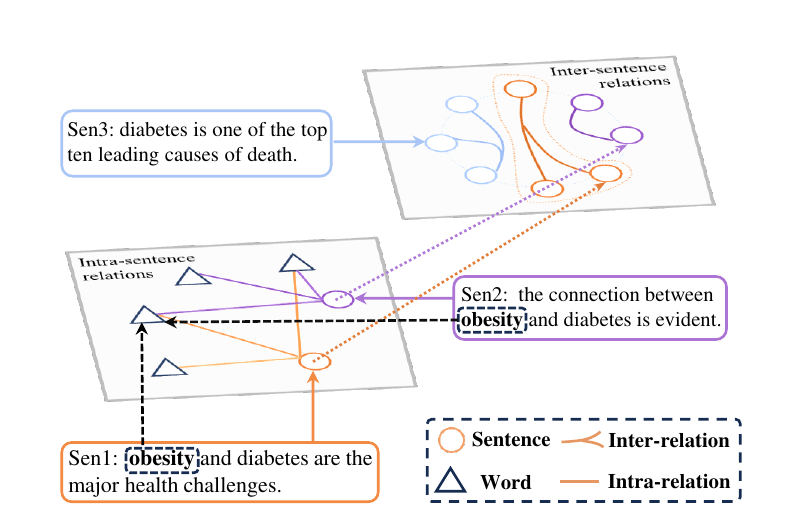}
\caption{An illustration of modeling an input document from local and global perspectives. Triangles and circles represent words and sentences in the original document respectively.}
\label{pic1}
\end{figure}The considerable length of the text hinders sequential models from capturing both long-range dependencies across sentences and intra-sentence relations simultaneously \citep{wang2020heterogeneous}.Moreover, the extended context exceeds the input limits of the Transformer-based model \citep{vaswani2017attention} due to the quadratic computational complexity of self-attention.

Recently, the application of large language models (LLM) such as ChatGPT to text summarization tasks has gained significant interest and attracted widespread attention. A recent study by \citep{zhang2023extractive} evaluated the performance of ChatGPT on extractive summarization and further enhanced its performance through in-context learning and chain-of-thought. Another study \citep{ravaut2023position} conducted experiments on abstractive summarization using various LLMs on a variety of datasets that included long inputs. While the use of LLMs in text summarization tasks has demonstrated exciting potential, there are still several limitations that have not been addressed. The most important of these is the phenomenon of lost-in-the-middle \citep{liu2023lost,ravaut2023position}, where LLMs ignore information in the middle and pay more attention to the context at the beginning and end. This bias raises concerns especially in summarization tasks where important text may be scattered throughout the document \citep{wu2023less}. Additionally, as the input length increases, even on explicitly long-context models, the model's performance gradually declines \citep{liu2023lost}. 

As a result, researchers have turned to graph neural networks to model long-distance relations. They represent a document as a graph and update node representations in the graph using message passing. These works use different methods to construct a graph from documents, such as using sentence similarity as edge weights to model cross-sentence relations \citep{zheng2019sentence}. Another popular approach is to construct a word-document heterogeneous graph \citep{wang2020heterogeneous}, using words as intermediate connecting sentences. \citet{phan2022hetergraphlongsum} further added passage nodes to the heterogeneous graph to enhance the semantic information. \citet{zhang2022hegel} proposed a hypergraph transformer to capture high-order cross-sentence relations.

Despite the impressive success of these approaches, we observe that the current work still lacks a comprehensive consideration on relational modeling. More specifically, two limitations are mentioned: (1) Most of the existing approaches focus on modeling intra-sentence relations but often overlook cross-sentence high-order relations. Inter-sentence connections may not only be pairwise but could also involve triplets or higher-order relations \citep{ding2020more}. In the hierarchical discourse structure of scientific documents, sentences within the same section often express the same main idea. It is difficult to fully understand the content of a document by merely considering intra-sentence and cross-sentence relations in pairwise. (2) These approaches rely on updating relations at different levels simultaneously but ignore the hierarchical structure of scientific documents. Sentences are composed of words and, in turn, contribute to forming sections. By understanding the meaning of individual tokens, we get the meaning of the sentence and thus the content of the section. Therefore, bottom-to-top structured modeling is crucial to understand the content of the document. 

To address the above challenges, we propose HAESum (\textbf{H}ierarchical \textbf{A}ttention Graph for \textbf{E}xtractive Document \textbf{Sum}marization), a method that leverages a graph neural network model to fully explore hierarchical structural information in scientific documents. HAESum first constructs a local heterogeneous graph of word-sentence and updates sentence representations at the intra-sentence level. The local sentence representations are then fed into a novel hypergraph self-attention layer to further update and learn the cross-sentence sentence representations through a self-attention mechanism that fully captures the relations between nodes and edges. Figure \ref{pic1} is an illustration showing the modeling of local and global context information from a hierarchical point of view, and the resulting representations contain both local and global hierarchical information. We validate HAESum with extensive experiments on two benchmark datasets and the experimental results demonstrate the effectiveness of our proposed method. In particular, we highlight our main contributions as follows:

\textbf{(\textrm{i})} We introduce a novel graph-based model utilizing the hierarchical structure of scientific documents for modeling. In contrast to simultaneously updating nodes in the graph, we learn intra-sentence and inter-sentence relations separately from both local and global perspectives. To the best of our knowledge, we are the first approach to hierarchical modeling using different graphs on this task.

\textbf{(\textrm{ii})} We propose a novel hypergraph self-attention layer that utilizes the self-attention mechanism to further aggregate high-order sentence representations. Moreover, our approach does not rely on pre-trained models as encoders, making it easily applicable to other low-resource languages.

\textbf{(\textrm{iii})} We validate our model on two benchmark datasets, and the experimental results demonstrate the effectiveness of our approach against strong baselines.
\section{Related Work}
\begin{figure*}[htbp]
\centering
    \includegraphics[scale=0.6]{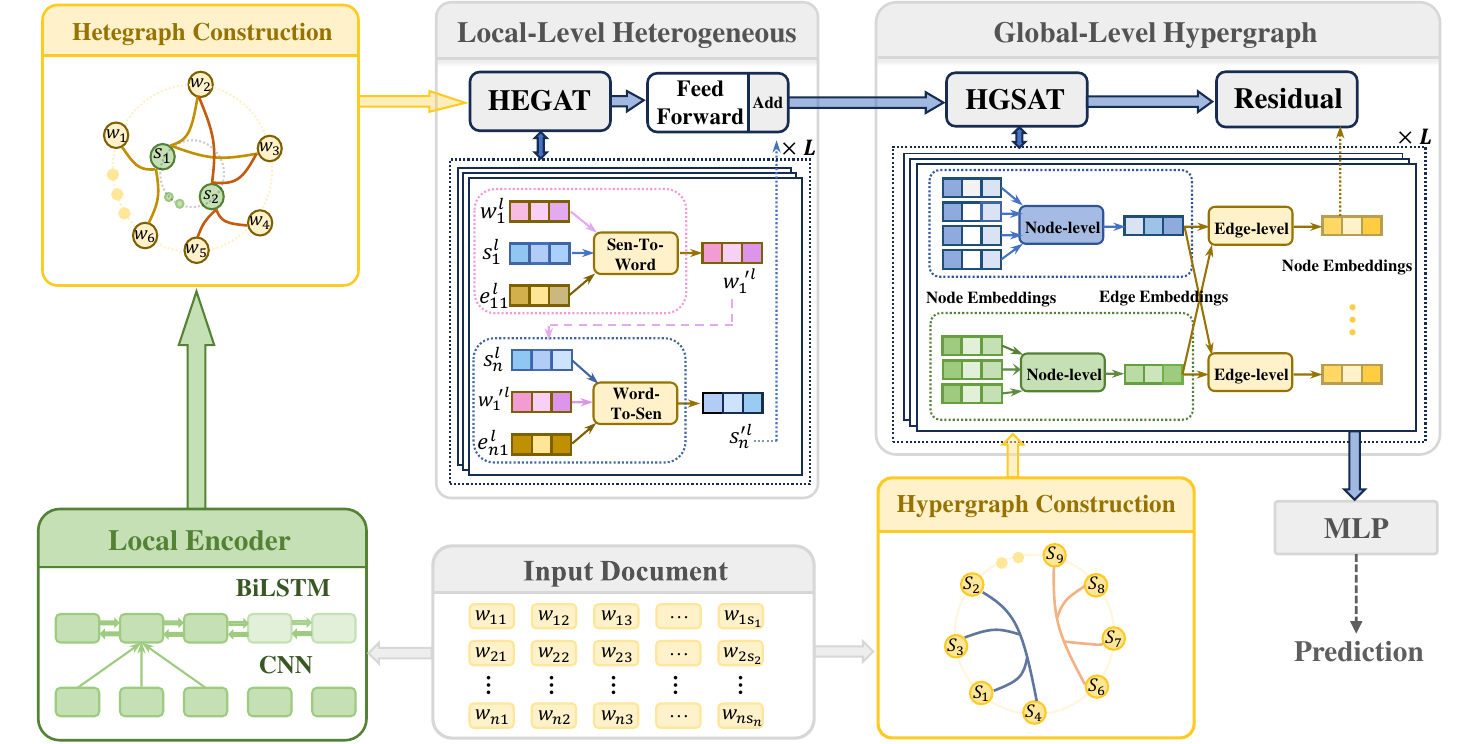}
\caption{Overview of the proposed HAESum framework. We first build a local-level heterogeneous graph (HEGAT) for the input document and apply message passing to iteratively update the representation in two stages: sentence-to-word and word-to-sentence. The obtained sentence representations are then fed into the hypergraph self-attention layer (HGSAT) to obtain the global representations and used for the final sentence selection.}
\label{pic2}
\end{figure*}

\subsection{Scientific Paper Summarization}
Scientific document summarization has been a hot topic due to the challenges of modeling long texts \citep{frermann2019inducing}. \citet{cohan2018discourse} introduced two benchmark datasets for long documents, Arxiv and PubMed, and employed a hierarchical encoder and discourse-aware decoder for the document summarization task. \citet{cui2021sliding} proposed a sliding selector network accompanied by dynamic memory to alleviate information loss between context segments. \citet{gu2021memsum} presented a reinforcement learning-based method that achieved impressive performance by considering the extraction history at each time step. Recently, \citet{ruan2022histruct+} proposed a method to inject explicit hierarchical structural information such as section titles and sentence positions into a pre-trained model to further improve the performance and interpretability.

\subsection{Graph based Summarization}
Graph neural networks have been widely used for extractive summarization due to their flexibility and scalability. \citet{dong2020discourse} proposed an unsupervised graph-based model that combines both sentence similarity and hierarchical discourse structure to rank sentences. \citet{cui2020enhancing} injected latent topic information into graph neural networks to further improve performance. \citet{wang2020heterogeneous} constructed a word-document heterogeneous graph using word nodes as intermediate to connect sentences. \citet{zhang2022hegel} proposed a hypergraph transformer to model long-distance dependency while emphasizing the importance of high-order inter-sentence relations in extraction summarization. Our paper follows this line of work, but the main difference is that our approach combines both intra-sentence relations and high-order cross-sentence relations and efficiently leverages the hierarchical discourse structure of scientific documents to learn sentence representations that incorporate both local and global information.
\section{Method}
Given an arbitrary document $ D=\{s_1,s_2,...,s_n\} $ consisting of $ n $ sentences, each sentence consists of $ m $ words $ s_i=\{w_{i1},w_{i2},...,w_{im}\} $. The goal of extractive summarization is to predict labels $ y_i \in \{0,1\} $ for all sentences, where $ y_i=1 $ indicates that the current sentence should be included in the summary. The overall structure of HAESum is shown in Figure \ref{pic2}.
\subsection{Local-level Heterogeneous Graph}
As the lowest level of the hierarchical structure, in this section, we will first introduce how to capture local intra-sentence relations between sentences and their corresponding words using a heterogeneous graph. We will start by explaining how to construct the heterogeneous graph and initialize it, followed by detailing how to use a heterogeneous self-attention layer to update node representations. Finally, we will feed the updated sentence node representations into the next module.
\subsubsection{Graph Construction}
Given an input document $D$, we first construct a heterogeneous graph $G=\{V, E\}$, where $V$ represents a set of nodes and $E$ represents edges between nodes. In order to utilize the natural hierarchy between words and sentences of a document, the nodes can be defined as $V=V_w\cup V_s$, where $V_w=\{w_1,w_2...,w_n\}$ denotes $n$ different words in the document, and $V_s=\{s_1,s_2,...,s_m\}$ denotes the $m$ sentences in the document. The edges are defined as $E=\{e_{11},e_{12},...,e_{mn}\}$, where $e_{ij}$ is a real-valued edge weight that denotes the cross-connection between a sentence node $i$ and a word node $j$ contained by it.
\subsubsection{Graph Initializers}
Let $X_w \in R^{|V_w|\times d_w}$, $X_s \in R^{|V_s|\times d_s }$ denote the feature matrices of the input word and sentence respectively. $d_w$ and $d_s$ correspond to the feature dimensions of words and sentences, respectively. We first use Glove \citep{pennington2014glove} to initialize word representations. Instead of using pre-trained model as a sentence encoder, we first use CNN \citep{lecun1998gradient} with different kernel sizes to get the n-gram feature $S^C$ of the sentence followed by using BiLSTM \citep{hochreiter1997long} to obtain the sentence-level feature $S^B$.The features obtained from CNN and BiLSTM are concatenated as initialized sentence representations $X_S=\text{Cat}(S^C, S^B)$.
\subsubsection{Heterogeneous Attention Modules}
Following the previous work \citep{wang2020heterogeneous}, we employ the heterogeneous graph attention layer for node representations updating. Specifically, when a node $v_i$ aggregates information from its neighbours, the attention coefficient $\alpha_{ij}$ for node $v_j$ is computed as follows:
\begin{equation}
    z_{ij}=\text{LeakyReLU}(W_a[W_sh_i\Vert W_kh_j];e_{ij})
\end{equation}
\begin{equation}
    \alpha_{ij}= \frac{\text{exp}(z_{ij})}{\sum_{l\in \mathcal{N}}\text{exp}(z_{il})}
\end{equation}
where $W_a$, $W_s$, $W_k$ are trainable weights. $\Vert$ denotes concatenation. We also inject the edge features $e_{ij}$ into the attention mechanism for computation.

We also add multi-head attention and Feed-Forward layer (FFN) \citep{vaswani2017attention} to further improve the performance. The final representation $u_i^{'}$ of node $v_i$ is then obtained as follows:
\begin{equation}
    u_i=\Vert_{k=1}^K\sigma(\sum_{j\in \mathcal{N}}\alpha_{ij}^kW^kh_i)
\end{equation}
\begin{equation}
    u_l^{'}=\text{FFN}(u_i)+h_i
\end{equation}
We begin by aggregating the sentence nodes around the word to update word representations. Subsequently, we utilize the updated word representations to further update the sentence representations.

In this section, we use the local heterogeneous graph to learn the intra-sentence relations at the lowest level of the document hierarchy.

\subsection{Global-level Hypergraph}
In this section, we first introduce how to construct a hypergraph. Subsequently, we present a novel hypergraph self-attention layer designed to fully capture high-order global inter-sentence relations. Finally, the resulting sentence representations are used to decide whether to include them in the summary.
\subsubsection{Hypergraph Construction}
A hypergraph is defined as $G=\{V,E\}$, where $V=\{v_1,v_2,...,v_n\}$ represents a set of nodes and $E=\{e_1,e_2...,e_n\}$ represents hyperedges in the graph. Unlike edges in regular graphs, hyperedges can connect two or more nodes and thus represent multivariate relations. A hypergraph is typically represented by its incidence matrix $\textbf{H} \in R^{n\times m}$ :
\begin{equation}
\label{eq:example}
\textbf{H}_{ij} = 
\begin{aligned}
    \left\{
    \begin{array}{l}
        1, \text{if} \ v_i \in e_j \\
        0, \text{if} \ v_i \notin e_j
    \end{array}
    \right.
\end{aligned}
\end{equation}
where $v_i\in V$, $e_j\in E$ and if the hyperedge $e_j$ connects node $v_i$ there is $v_i\in e_j$.

We denote a sentence $s_i$ in a document $D=\{s_1,s_2,...,s_n\}$ as a node $v_i$ in the hypergraph. In order to capture global higher-order inter-sentence relations, we consider creating section hyperedges for each part \citep{suppe1998structure}.  A hyperedge $e_j$ will be created if a set of child nodes $V_j \in V$ belongs to the same section in the document. The node representations in the hypergraph are initialized to the output of the previous module.

The initialized node features $H_{sen}=\{h_1,h_2,...,h_n\}\in R^{n\times d}$ and incidence matrix $H$ will be fed into the hypergraph self-attention network to learn effective sentence representations.
\subsubsection{Hypergraph Self-Attention Modules}
Hypergraph attention networks (HGAT) are designed to learn node representations using a mutual attention mechanism. This mutual attention mechanism divides the computational process into two steps, i.e., node aggregation and hyperedge aggregation. First the hyperedge representations are updated with node information. Subsequently, the hyperedge information is fused back to the nodes from hyperedges.

The HGAT has mainly been implemented based on graph attention mechanism \citep{velivckovic2017graph}, such as HyperGAT \citep{ding2020more}. However, this attention mechanism employs the same weight matrix for different types of nodes and hyperedges information and could not fully exploit the relations between nodes and hyperedges, which prevents the model from capturing higher-order cross-sentence relations \citep{fan2021heterogeneous}.

To address the limitations of HGAT, we propose the hypergraph self-attention layer. Inspired by the success of Transformer \citep{vaswani2017attention} in textual representation and graph learning \citep{ying2021transformers}, we use the self-attention mechanism to fully explore the relations between nodes and hyperedges. The entire structure we propose is described below.

\noindent
\textbf{Node-level Attention} \, To solve the problem of initializing the hyperedge features, we first encode hyperedge representations from node aggregation information using node-level attention. Given node features $H_{sen}^{l-1}=\{h_1^{l-1},h_2^{l-1},...,h_n^{l-1}\}$ and incidence matrix, hyperedge representations $\{f_1^l,f_2^l...,f_m^l\}$ can be computed as follows:
\begin{equation}
    f_j^l=\text{LeakyReLU}(\sum_{s_k \in e_j}\alpha_{jk}W_nh_k^{l-1})
\end{equation}
\begin{equation}
    \alpha_{jk}= \frac{\text{exp}(W_h^\text{T}u_k)}{\sum_{s_l \in e_j}\text{exp}(W_h^\text{T}u_l)}
\end{equation}

\begin{equation}
    u_k=\text{LeakyReLU}(W_{p}h_k^{l-1})
\end{equation}
where the superscript $l$ denotes the model layer. $W_n$, $W_h$, $W_p$ are trainable parameters. $\alpha_{jk}$ is the attention coefficient of node $s_k$ in the hyperedge $e_j$. Through the node-level attention mechanism, we initialize the hyperedge representation.

\noindent
\textbf{Edge-level Attention} \, As an inverse procedure, the self-attention mechanism is applied to compute the importance scores to highlight the hyperedges that are more critical for the next layer of node representation $v_i$. Given the node feature matrix $H_{sen}^{l-1}$ and the hyperedge feature matrix $F_{edge}^l$, similar to the self-attention mechanism we compute the output matrix as follows:
\begin{equation}
    \begin{split}
        Q_{sen}^{l-1} &=W_qH_{sen}^{l-1} \\
        K_{edge}^l &=W_kF_{edge}^l \\
        V_{edge}^l &=W_vF_{edge}^l \\
    \end{split}
\end{equation}
\begin{equation}
    \text{Att}(H, F) = \text{softmax}(\frac{Q_{sen}^{l-1}{K_{edge}^l}^\text{T}}{\sqrt{d_k}})V_{edge}^l
\end{equation}
where $W_q$,$W_k$,$W_v$ are trainable parameters. $d_k$ is the feature dimension of the hidden layer.
$\text{Att}()$ represents the self-attention mechanism.

After obtaining the enhanced node representations $H_{sen}^l$ using the hypergraph self-attention layer, we applied a feature fusion layer to generate the final representations $H_{sen}^{'l}$, which can be represented by the formula:
\begin{equation}
    H_{sen}^{'l} = \text{LeakyReLU}(W_1H_{sen}^{l-1}\Vert W_2H_{sen}^l)
\end{equation}
$\Vert$ denotes concatenation. Fusing hyperedge information and node information, we obtain a semantic representation of sentence nodes.
\subsection{Opimization}
After passing L hypergraph self-attention layers, we obtain the representations of sentences $H_{sen}=\{h_1,h_2,...,h_n\} \in R^{n\times d}$.  We then add a multi-layer perceptron (MLP) followed by a LayerNorm layer and obtain a score $\hat{y}_i$, indicating whether it will be selected as a summary. Formally, the prediction score for a sentence node $s_i$ is computed as follows:
\begin{equation}
    \hat{y}_i = W_o(\text{LayerNorm}(W_ph_i))
\end{equation}
where $W_o$,$W_p$ are trainable parameters. 

Finally, the output sentence scores $\hat{y}_i$ are optimized with the true labels $y_i$ by binary cross-entropy loss:
\begin{equation}
    L=\frac{1}{N}\sum_{i=1}^Ny_i\text{log}\hat{y}_i + (1-y_i)\text{log}(1-\hat{y}_i)
\end{equation}
where $N$ denotes the number of sentences in the document.

\begin{table}[]
\resizebox{\columnwidth}{!}{%
\begin{tabular}{c|ccc|cc}
\hline
\multirow{2}{*}{\textbf{Datasets}} & \multicolumn{3}{c|}{\textbf{Document}}        & \multirow{2}{*}{\textbf{\begin{tabular}[c]{@{}c@{}}Avg.\\ Doc.\end{tabular}}} & \multirow{2}{*}{\textbf{\begin{tabular}[c]{@{}c@{}}Avg.\\ Token.\end{tabular}}} \\
                                   & \textbf{Train} & \textbf{Val} & \textbf{Test} &                                                                               &                                                                                 \\ \hline
Arxiv                              & 202703         & 6436         & 6439          & 4938                                                                          & 220                                                                             \\ \hline
PubMed                             & 116669         & 6630         & 6657          & 3016                                                                          & 203                                                                             \\ \hline
\end{tabular}%
}
\caption{Statistics of Arxiv and PubMed datasets.}
\label{tab1}
\end{table}

\section{Experiment}
\subsection{Experiment setup}
We validate our proposed model on two scientific document datasets and compare it to the strong baselines. In the following, we start with the details of the datasets.

\noindent
\textbf{Datasets} \, We perform extensive experiments on two benchmark datasets: Arxiv and PubMed \citep{cohan2018discourse}. Arxiv is a long document dataset containing different scientific domains. PubMed contains articles in the biomedical domain. We use the original train, validation, and testing splits as in \citep{cohan2018discourse}. Detailed statistics for the two benchmark datasets are shown in Table \ref{tab1}.
\begin{table*}[]
\centering
\resizebox{0.95\textwidth}{!}{%
\begin{tabular}{cll|clclcl|clclcl}
\hline
\multicolumn{3}{c|}{\multirow{2}{*}{Models}} &
  \multicolumn{6}{c|}{PubMed} &
  \multicolumn{6}{c}{Arxiv} \\ \cline{4-15} 
\multicolumn{3}{c|}{} &
  \multicolumn{2}{c}{ROUGE-1} &
  \multicolumn{2}{c}{ROUGE-2} &
  \multicolumn{2}{c|}{ROUGE-L} &
  \multicolumn{2}{c}{ROUGE-1} &
  \multicolumn{2}{c}{ROUGE-2} &
  \multicolumn{2}{c}{ROUGE-L} \\ \hline
\multicolumn{3}{c|}{Oracle} &
  \multicolumn{2}{c}{55.05} &
  \multicolumn{2}{c}{27.48} &
  \multicolumn{2}{c|}{49.11} &
  \multicolumn{2}{c}{53.88} &
  \multicolumn{2}{c}{23.05} &
  \multicolumn{2}{c}{46.54} \\
\multicolumn{3}{c|}{PacSum} &
  \multicolumn{2}{c}{39.79} &
  \multicolumn{2}{c}{14.00} &
  \multicolumn{2}{c|}{36.09} &
  \multicolumn{2}{c}{38.57} &
  \multicolumn{2}{c}{10.93} &
  \multicolumn{2}{c}{34.33} \\
\multicolumn{3}{c|}{HIPORANK} &
  \multicolumn{2}{c}{43.58} &
  \multicolumn{2}{c}{17.00} &
  \multicolumn{2}{c|}{39.31} &
  \multicolumn{2}{c}{39.34} &
  \multicolumn{2}{c}{12.56} &
  \multicolumn{2}{c}{34.89} \\
\multicolumn{3}{c|}{FAR} &
  \multicolumn{2}{c}{41.98} &
  \multicolumn{2}{c}{15.66} &
  \multicolumn{2}{c|}{37.58} &
  \multicolumn{2}{c}{40.92} &
  \multicolumn{2}{c}{13.75} &
  \multicolumn{2}{c}{35.56} \\ \hline
\multicolumn{3}{c|}{ExtSum-LG} &
  \multicolumn{2}{c}{44.85} &
  \multicolumn{2}{c}{19.70} &
  \multicolumn{2}{c|}{31.43} &
  \multicolumn{2}{c}{43.62} &
  \multicolumn{2}{c}{17.36} &
  \multicolumn{2}{c}{29.14} \\
\multicolumn{3}{c|}{Topic-GraphSum} &
  \multicolumn{2}{c}{45.95} &
  \multicolumn{2}{c}{20.81} &
  \multicolumn{2}{c|}{33.97} &
  \multicolumn{2}{c}{44.03} &
  \multicolumn{2}{c}{18.52} &
  \multicolumn{2}{c}{32.41} \\
\multicolumn{3}{c|}{SSN-DM} &
  \multicolumn{2}{c}{46.73} &
  \multicolumn{2}{c}{21.00} &
  \multicolumn{2}{c|}{34.10} &
  \multicolumn{2}{c}{45.03} &
  \multicolumn{2}{c}{19.03} &
  \multicolumn{2}{c}{32.58} \\
\multicolumn{3}{c|}{HEGEL} &
  \multicolumn{2}{c}{47.13} &
  \multicolumn{2}{c}{21.00} &
  \multicolumn{2}{c|}{42.18} &
  \multicolumn{2}{c}{46.41} &
  \multicolumn{2}{c}{18.17} &
  \multicolumn{2}{c}{39.89} \\
\multicolumn{3}{c|}{MTGNN} &
  \multicolumn{2}{c}{48.42} &
  \multicolumn{2}{c}{22.26} &
  \multicolumn{2}{c|}{43.66} &
  \multicolumn{2}{c}{46.39} &
  \multicolumn{2}{c}{18.58} &
  \multicolumn{2}{c}{40.50} \\
\multicolumn{3}{c|}{HiStruct+} &
  \multicolumn{2}{c}{46.59} &
  \multicolumn{2}{c}{20.39} &
  \multicolumn{2}{c|}{42.11} &
  \multicolumn{2}{c}{45.22} &
  \multicolumn{2}{c}{17.67} &
  \multicolumn{2}{c}{40.16} \\
\multicolumn{3}{c|}{CHANGES} &
  \multicolumn{2}{c}{46.43} &
  \multicolumn{2}{c}{21.17} &
  \multicolumn{2}{c|}{41.58} &
  \multicolumn{2}{c}{45.61} &
  \multicolumn{2}{c}{18.02} &
  \multicolumn{2}{c}{40.06} \\ \hline
\multicolumn{3}{c|}{TLM-I+E} &
  \multicolumn{2}{c}{42.13} &
  \multicolumn{2}{c}{16.27} &
  \multicolumn{2}{c|}{39.21} &
  \multicolumn{2}{c}{41.62} &
  \multicolumn{2}{c}{14.69} &
  \multicolumn{2}{c}{38.03} \\
\multicolumn{3}{c|}{PEGASUS} &
  \multicolumn{2}{c}{45.49} &
  \multicolumn{2}{c}{19.90} &
  \multicolumn{2}{c|}{42.42} &
  \multicolumn{2}{c}{44.70} &
  \multicolumn{2}{c}{17.27} &
  \multicolumn{2}{c}{25.80} \\
\multicolumn{3}{c|}{BigBird} &
  \multicolumn{2}{c}{46.32} &
  \multicolumn{2}{c}{20.65} &
  \multicolumn{2}{c|}{42.33} &
  \multicolumn{2}{c}{46.63} &
  \multicolumn{2}{c}{19.02} &
  \multicolumn{2}{c}{\textbf{41.77}} \\
\multicolumn{3}{c|}{Dancer} &
  \multicolumn{2}{c}{46.34} &
  \multicolumn{2}{c}{19.97} &
  \multicolumn{2}{c|}{42.42} &
  \multicolumn{2}{c}{45.01} &
  \multicolumn{2}{c}{17.60} &
  \multicolumn{2}{c}{40.56} \\ \hline
  \multicolumn{3}{c|}{ChatGLM3-6B-32k}            & \multicolumn{2}{c}{40.95}            & \multicolumn{2}{c}{15.79}            & \multicolumn{2}{c|}{37.09}            & \multicolumn{2}{c}{39.81}            & \multicolumn{2}{c}{14.14}            & \multicolumn{2}{c}{35.36}            \\ \hline
\multicolumn{3}{c|}{\textbf{HAESum (ours)}} &
  \multicolumn{2}{c}{\textbf{48.77}} &
  \multicolumn{2}{c}{\textbf{22.44}} &
  \multicolumn{2}{c|}{\textbf{43.83}} &
  \multicolumn{2}{c}{\textbf{47.24}} &
  \multicolumn{2}{c}{\textbf{19.44}} &
  \multicolumn{2}{c}{41.34} \\ \hline
\end{tabular}%
}
\caption{Experimental Results on PubMed and Arxiv datasets. We report ROUGE scores from the original papers if available, or scores from \citep{xiao2019extractive} otherwise.}
\label{tab2}
\end{table*}

\noindent
\textbf{Compared Baselines} \, We make a systematic comparison with recent approaches in this area. We categorize these methods into the following four types:

\begin{itemize}
\setlength{\itemsep}{1pt}
\setlength{\parsep}{0pt}
\setlength{\parskip}{0pt}
\item Unsupervised methods: graph-based models PacSum \citep{zheng2019sentence}, HIPORANK \citep{dong2020discourse}, FAR \citep{liang2021improving}.
\item Neural extractive model: Seq2Seq-based models HiStruct+ \citep{ruan2022histruct+}; local and global context model ExtSum-LG \citep{xiao2019extractive}; graph-based models Topic-GraphSum \citep{cui2020enhancing}, SSN-DM \citep{cui2021sliding}, HEGEL \citep{zhang2022hegel}, MTGNN \citep{doan2022multi}, CHANGES \citep{zhang2023contrastive}.
\item Neural abstractive model: encoder-decoder based Model TLM-I+E \citep{pilault2020extractive}, PEGASUS \citep{zhang2020pegasus} , BigBird \citep{zaheer2020big}, divide-and-conquer approach Dancer \citep{gidiotis2020divide}.
\item Large language model: ChatGLM3-6k-32k \citep{zeng2022glm}.
More details on the evaluation of the large language model can be found in Appendix \ref{sec:llms}.
\end{itemize}
\subsection{Implementation Details}
Regarding the encoding of word nodes, the vocabulary size is 50000 and the word embedding is initialized with a dimension of 300 using the Glove pre-trained model\citep{pennington2014glove}. The feature dimensions of sentence nodes and edges in the heterogeneous graph are set to 64 and 50, respectively. The hyperedge feature dimension is 64. We set the maximum sentence length of each document to 200 and the maximum number of words per sentence to 100. In our experiments, we stacked two layers of heterogeneous graph attention modules (HEGAT) and hypergraph self-attention modules (HSAGT). The multi-head of the HEGAT layer is set to 8 and 6, respectively. 

The model is optimized using the Adam optimizer \citep{loshchilov2017decoupled} with a learning rate of 0.0001 and a dropout rate of 0.1. We train the model on an RTX A6000 GPU with 48GB of memory for 12 epochs. The training process stops if the validation set loss does not decrease three times. The training time for one epoch on the PubMed dataset is 3 hours, while on the Arxiv dataset, it is 6 hours.

We use a greedy search algorithm similar to \citep{nallapati2017summarunner} to select sentences from documents as the gold extractive summaries (Oracle). Following previous work, we use ROUGE \citep{lin2003automatic} to evaluate the quality of summaries. We use ROUGE-1/2 to measure summary informativeness and ROUGE-L to measure the fluency of the summary.

\subsection{Experiment Results}
Table \ref{tab2} shows the comparison between our model HAESum and the baseline model on PubMed and Arxiv datasets. The first block covers the ground truth ORACLE and unsupervised methods for extractive summarization. The second block covers state-of-the-art supervised extractive baselines. The third block reports abstractive methods.

Based on the results, we find that HIPORANK \citep{dong2020discourse} achieves strong performance on graph-based unsupervised modeling. Compared to other unsupervised methods, HIPORANK adds section information, which demonstrates the effectiveness and importance of taking the natural hierarchical structure of scientific documents into account when modeling cross-sentence relations.

\begin{table}[]
\resizebox{\columnwidth}{!}{%
\begin{tabular}{cccc}
\hline
\multicolumn{1}{c|}{Model}              & ROUGE-1        & ROUGE-2        & ROUGE-L        \\ \hline
\multicolumn{4}{c}{PubMed}                                                                 \\ \hline
\multicolumn{1}{c|}{\textbf{HAESum}}    & \textbf{48.77} & \textbf{22.44} & \textbf{43.83} \\
\multicolumn{1}{c|}{w/o Heterogeneous}  & 47.45          & 21.12          & 42.56          \\
\multicolumn{1}{c|}{w/o HyperAttention} & 47.60          & 21.43          & 42.78          \\ \hline
\multicolumn{4}{c}{Arxiv}                                                                  \\ \hline
\multicolumn{1}{c|}{\textbf{HAESum}}    & \textbf{47.24} & \textbf{19.44} & \textbf{41.34} \\
\multicolumn{1}{c|}{w/o Heterogeneous}  & 46.91          & 19.22          & 41.03          \\
\multicolumn{1}{c|}{w/o HyperAttention} & 46.75          & 19.01          & 40.91          \\ \hline
\end{tabular}%
}
\caption{Ablation study results on PubMed and Arxiv datasets.}
\label{tab3}
\end{table}

In the extractive baseline, MTGNN \citep{doan2022multi} achieves state-of-art performance, MTGNN considers more intra-sentence level modeling, which shows the necessity of modeling from low-level structure. HEGEL \citep{zhang2022hegel} is the most similar approach to ours. HEGEL injects external information such as keywords and topics into the model and models higher-order cross-sentence relations through a hypergraph transformer to achieve a competitive performance. However, compared to MTGNN, HEGEL does not consider low-level intra-sentence relations, which proves the necessity of considering and modeling hierarchical structure. Interestingly, CHANGES \citep{zhang2023contrastive} achieves equally impressive results in hierarchical modeling by considering high-level intra-section and inter-section relations, further confirming the importance of hierarchical modeling. Among the extractive methods, the transformer-based HiStruct+ \citep{ruan2022histruct+} shows a competitive performance, which demonstrates the effectiveness of the self-attention mechanism. HiStruct+ also incorporates the inherent hierarchical structure into the pre-trained language models to achieve strong performance. In addition, the extractive approaches largely outperform the abstractive approaches, which may be due to the fact that long input is more challenging for the decoding process of the abstractive models.

Through the table, the results of using the large language model are not satisfactory compared to our proposed method. By analyzing the output of the large language model, the model sometimes incorrectly outputs content from other languages and also occasionally outputs duplicate content. In addition, the model sometimes misinterprets extractive summarization as abstractive summarization.

According to the experimental results, our model HAESum outperforms all extractive and abstractive strong baselines. In particular, our model neither requires injection of external knowledge (e.g., topics and keywords \citep{zhang2022hegel}) to enhance global information nor pre-trained model's (e.g., BERT \citep{devlin2018bert}) knowledge \citep{doan2022multi}. The outstanding performance of HAESum demonstrates the importance of hierarchical modeling of local intra-sentence relations and global inter-sentence relations.

\begin{table}[]
\resizebox{\columnwidth}{!}{%
\begin{tabular}{c|ccc}
\hline
Method                      & ROUGE-1        & ROUGE-2        & ROUGE-L        \\ \hline
\textbf{Hierarchical(Ours)} & \textbf{48.77} & \textbf{22.44} & \textbf{43.83} \\ 
Parallelization             & 48.36          & 22.03          & 43.36          \\ \hline
\end{tabular}%
}
\caption{Different ways of updating sentence representations on PubMed dataset.}
\label{tab4}
\end{table}

\begin{figure*}[htbp]
\centering
    \includegraphics[scale=0.6]{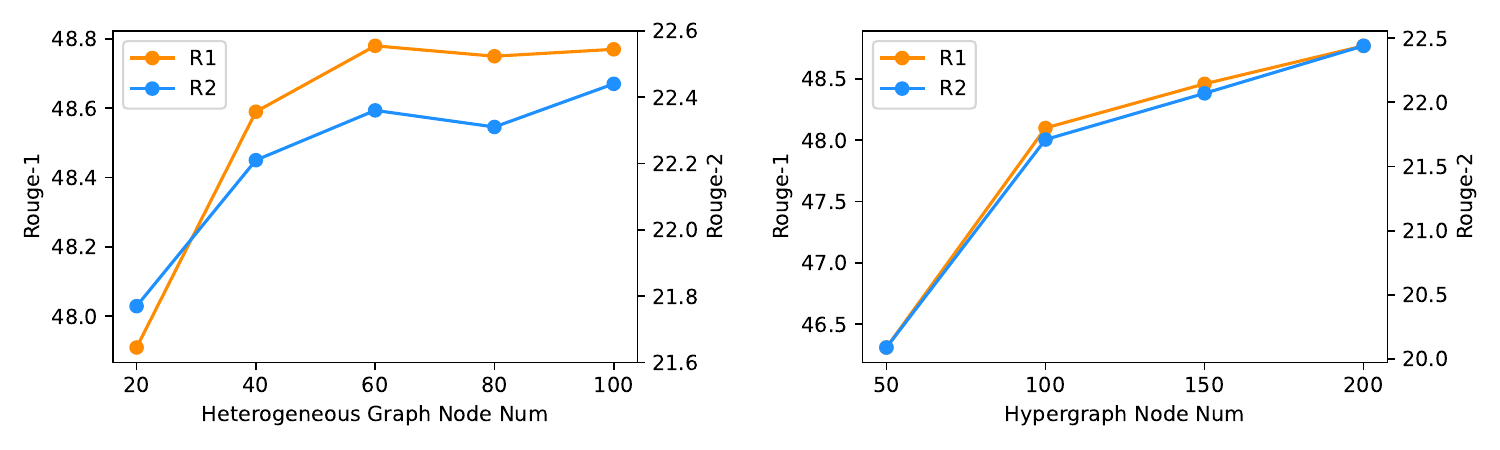}
\caption{ROUGE-1,2 performance of HAESum with different number of graph nodes on PubMed dataset.}
\label{pic3}
\end{figure*}

\section{Analysis}
\subsection{Ablation Study}
We first analyze the effect of different components of HAESum in Table \ref{tab3}. The second row shows that removing the heterogeneous graph part represents not learning intra-sentence relations. The third row removes the hypergraph component, representing the absence of learning higher-order cross-sentence relations. As shown in table \ref{tab3}, removing either part hurts the model performance, which indicates that learning both local intra-sentence relations and global higher-order cross-sentence relations is necessary for scientific document summarization.

Interestingly, these two components are almost equally important for modeling long documents. This indicates the importance of simultaneously modeling semantic aspects from diverse perspectives and hierarchical discourse structures in scientific documents.

\subsection{Performance Analysis}
\noindent
\textbf{Hierarchical discourse} \, We also analyze different update approaches for obtaining the final sentence representations in HAESum. As shown in Table \ref{tab4}, the second row represents our hierarchical updating. The third row represents parallel updating, where intra-sentence and inter-sentence relations are updated simultaneously, and the final sentence representations are concatenated. The superior performance of hierarchical updating over parallel updating once again emphasizes the critical importance of the bottom-to-top modeling sequence we propose for understanding the content of long documents.

\begin{table}[]
\resizebox{\columnwidth}{!}{%
\begin{tabular}{cccc}
\hline
\multicolumn{1}{c|}{Model}            & ROUGE-1        & ROUGE-2        & ROUGE-L        \\ \hline
\multicolumn{4}{c}{PubMed}                                                               \\ \hline
\multicolumn{1}{c|}{\textbf{with HSGAT (ours)}} & \textbf{48.77} & \textbf{22.44} & \textbf{43.83} \\
\multicolumn{1}{c|}{with HGAT}        & 48.64          & 22.25          & 43.64          \\ \hline
\multicolumn{4}{c}{Arxiv}                                                                \\ \hline
\multicolumn{1}{c|}{\textbf{with HSGAT (ours)}} & \textbf{47.24} & \textbf{19.44} & \textbf{41.34} \\
\multicolumn{1}{c|}{with HGAT}        & 47.08          & 19.26          & 41.18        \\ \hline
\end{tabular}%
}
\caption{Different attention mechanism results on PubMed and Arxiv datasets.}
\label{tab5}
\end{table}

\noindent
\textbf{Attention mechanism} \, We then analyze the performance of our proposed novel hypergraph self-attention layer and hypergraph attention network (HGAT). As shown in Table \ref{tab5}, our hypergraph self-attention layer outperforms HGAT \citep{ding2020more}. We speculate that the main reason is the utilization of the self-attention mechanism and different weight matrices, which fully exploit relations between nodes and edges, thereby enhancing the learning of high-order relations.

\noindent
\textbf{Hyperparameter sensitivity} \, In our experiments, we set the maximum input length for each sentence to be 100, and the maximum sentence length for each input document to be 200. We conduct an analysis of these two hyperparameters. In addition, more information about the distribution of the sentence lengths and the number of sentences in the document is presented in the Appendix \ref{sec:distri}. As shown in Figure \ref{pic3}, when the maximum number of tokens in each sentence is reduced from 100 to 60, the performance does not significantly decrease. This indicates that under this range of hyperparameter settings, the model has already processed most of the tokens in each sentence. However, as the length continues to decrease, the model's performance starts to decline, as the input length limits the capture of local intra-sentence relations.

Simultaneously, when the maximum number of sentences in a document is increased from 50 to 200, the model's performance continues to improve. This improvement is attributed to the consideration of more sentences, capturing more complex higher-order cross-sentence relations. However, persistently increasing this hyperparameter leads to significant computational consumption. Specifically, in future work, we intend to increase the maximum input sentences per document while minimizing computational consumption as much as possible.

\subsection{Case Study}
Here we provide an example of a summary output by HAESum, as shown in Table \ref{tab6}. The selected sentences are mainly from the same section and cover the entire document. This illustrates that HAESum can effectively learn both local intra-sentence and high-order inter-sentence relations, facilitating the selection of the most relevant sentences.

\begin{table}[h]
\resizebox{\columnwidth}{!}{%
\begin{tabular}{ll}
\hline
\multicolumn{2}{l}{\begin{tabular}[c]{@{}l@{}}\textbf{(Introduction)} It includes hidradenitis suppurativa acne congl-\\ obata dissecting cellulitis of the scalp and pilonidal sinus.\end{tabular}}                                                               \\
\multicolumn{2}{l}{\begin{tabular}[c]{@{}l@{}}\textbf{(Introduction)} Though each of these conditions are commonly\\ encountered on their own as a symptom complex follicular occ-\\ lusion tetrad has rarely been reported in the literature here.\end{tabular}} \\
\multicolumn{2}{l}{\begin{tabular}[c]{@{}l@{}}\textbf{(Introduction)} We present a case of hidradenitis suppurativa in\\ a 36-year-old male patient who also had the above mention-\\ ed associations.\end{tabular}}                                                 \\
\multicolumn{2}{l}{\begin{tabular}[c]{@{}l@{}}\textbf{(Case Report)} A 36-year-old male patient presented to us with\\ a history of recurrent boils since 18 years.\end{tabular}}                                                                                  \\
\multicolumn{2}{l}{\begin{tabular}[c]{@{}l@{}}\textbf{(Discussion)} Follicular occlusion tetrad is a condition that incl-\\ udes hidradenitis suppurativa (hs) acne conglobata dissecting\\ cellulitis of the scalp and pilonidal sinus.\end{tabular}}          \\ \hline
\end{tabular}%
}
\caption{An example output summary of our proposed model.}
\label{tab6}
\end{table}

\section{Conclusion}
This paper presents HAESum for scientific document summarization. HAESum employs a graph-based model to comprehensively learn local intra-sentence and high-order inter-sentence relations, utilizing the hierarchical discourse structure of scientific documents for modeling. The impressive performance of HAESum demonstrates the importance of simultaneously considering multiple perspectives of semantics and hierarchical structural information in modeling scientific documents.

\section*{Limitations}
Despite the outstanding performance of our HAESum, several limitations are acknowledged. Firstly, HAESum solely leverages intra-sentence and inter-sentence relations in scientific documents. We believe that incorporating other hierarchical discourse structures at different granularities, such as sentence-section information \citep{zhang2023contrastive} or dependency parsing trees, could further enhance model performance. Secondly,  although the context window sizes of large language models satisfy the input length of scientific documents, their performance on text summarization tasks, especially on long input texts, remains to be improved due to the loss-in-the-middle \citep{liu2023lost,ravaut2023position} problem. We consider this issue as a future work. Additionally, we focused on single document summarization. We believe that incorporating domain knowledge through citation networks and similar methods could further improve performance.

\section*{Acknowledgments}
This work is supported by the National Natural Science Foundation of China under Grant (No.61971057).

\bibliography{custom}

\clearpage
\begin{table*}[]
\centering
\resizebox{\textwidth}{!}{%
\begin{tabular}{c|lll|clclcl|clclcl}
\hline
\multirow{2}{*}{Models} &
  \multicolumn{3}{l|}{\multirow{2}{*}{\begin{tabular}[c]{@{}l@{}}\ Satisfy The\\ Input Length\end{tabular}}} &
  \multicolumn{6}{c|}{PubMed} &
  \multicolumn{6}{c}{Arxiv} \\ \cline{5-16} 
 &
  \multicolumn{3}{l|}{} &
  \multicolumn{2}{c}{ROUGE-1} &
  \multicolumn{2}{c}{ROUGE-2} &
  \multicolumn{2}{c|}{ROUGE-L} &
  \multicolumn{2}{c}{ROUGE-1} &
  \multicolumn{2}{c}{ROUGE-2} &
  \multicolumn{2}{c}{ROUGE-L} \\ \hline
ChatGPT &
  \multicolumn{3}{c|}{\ding{56}} &
  \multicolumn{2}{c}{-} &
  \multicolumn{2}{c}{-} &
  \multicolumn{2}{c|}{-} &
  \multicolumn{2}{c}{-} &
  \multicolumn{2}{c}{-} &
  \multicolumn{2}{c}{-} \\
LLaMa-7B &
  \multicolumn{3}{c|}{\ding{56}} &
  \multicolumn{2}{c}{-} &
  \multicolumn{2}{c}{-} &
  \multicolumn{2}{c|}{-} &
  \multicolumn{2}{c}{-} &
  \multicolumn{2}{c}{-} &
  \multicolumn{2}{c}{-} \\
ChatGLM3-6B &
  \multicolumn{3}{c|}{\ding{56}} &
  \multicolumn{2}{c}{-} &
  \multicolumn{2}{c}{-} &
  \multicolumn{2}{c|}{-} &
  \multicolumn{2}{c}{-} &
  \multicolumn{2}{c}{-} &
  \multicolumn{2}{c}{-} \\
ChatGLM3-6B-32k &
  \multicolumn{3}{c|}{\ding{52}} &
  \multicolumn{2}{c}{40.95} &
  \multicolumn{2}{c}{15.79} &
  \multicolumn{2}{c|}{37.09} &
  \multicolumn{2}{c}{39.81} &
  \multicolumn{2}{c}{14.14} &
  \multicolumn{2}{c}{35.36} \\ \hline
\textbf{HAESum(ours)} &
  \multicolumn{3}{c|}{\ding{52}} &
  \multicolumn{2}{c}{\textbf{48.77}} &
  \multicolumn{2}{c}{\textbf{22.44}} &
  \multicolumn{2}{c|}{\textbf{43.83}} &
  \multicolumn{2}{c}{\textbf{47.24}} &
  \multicolumn{2}{c}{\textbf{19.44}} &
  \multicolumn{2}{c}{\textbf{41.34}} \\ \hline
\end{tabular}%
}
\caption{Experimental results on large language models on two datasets}
\label{table7}
\end{table*}

\begin{table*}[]
\centering
\resizebox{\textwidth}{!}{%
\begin{tabular}{c|ccc|ccc}
\hline
\multirow{2}{*}{Model} & \multicolumn{3}{c|}{PubMed}         & \multicolumn{3}{c}{Arxiv}           \\ \cline{2-7} 
                       & Overall & Coverage & Non-Redundancy & Overall & Coverage & Non-Redundancy \\ \hline
ChatGLM3-6B-32K        & 2.52    & 2.51     & 2.41           & 2.48    & 2.45     & 2.29           \\
MTGNN                  & 1.73    & 1.74     & \textbf{1.67}  & 1.85    & 1.91     & 1.83           \\
\textbf{HAESum(Ours)} & \textbf{1.68} & \textbf{1.64} & 1.71 & \textbf{1.61} & \textbf{1.57} & \textbf{1.68} \\ \hline
\end{tabular}%
}
\caption{Average rank of human evaluation in terms of overall performance, coverage, and non-redundancy. Lower score is better.}
\label{tab8}
\end{table*}

\begin{table*}[]
\centering
\resizebox{\textwidth}{!}{%
\begin{tabular}{cccccc}
\hline
\multicolumn{6}{c}{The distribution of the sentence lengths}                    \\ \hline
(0, 20{]} & (20, 40{]}  & (40, 60{]}   & (60, 80{]}   & (80, 100{]}  & Over 100 \\ \hline
 29.63\% & 51.08\% & 12.82\% & 3.78\% & 1.38\% &  1.31\% \\ \hline
\multicolumn{6}{c}{The distribution of the number of sentences}                 \\ \hline
(0, 50{]} & (50, 100{]} & (100, 150{]} & (150, 200{]} & (200, 250{]} & Over 250 \\ \hline
28.09\%   & 40.83\%     & 19.36\%      & 7.59\%       & 2.57\%       & 1.56\%   \\ \hline
\end{tabular}%
}
\caption{The distribution of the sentence lengths and the number of sentences  in PubMed dataset}
\label{tab9}
\end{table*}

\appendix
\section{Appendix}
In this section, we give more details about the experiment.
\subsection{Evaluation on LLMs}
We tested different prompts and chose the best prompt. 
The \textbf{prompt} we used is: You are given a long scientific literature. Please read and choose no more than five sentences from the original scientific literature as a summary. Scientific literature:[\textit{Text Document}]. Now, select no more than five sentences from the original given scientific literature as a summary. Summary:[\textit{Output}].

The experimental results are shown in Table \ref{table7}, where we considered  a variety of possible large language models. However, in order to fulfill the requirement of inputting long texts, we chose ChatGLM3-6B-32K \citep{zeng2022glm} to evaluate the performance results on two datasets.

Through the table, the results of using the large language model are not satisfactory compared to our proposed method. By analyzing the output of the large language model, the model sometimes incorrectly outputs content from other languages and also occasionally outputs duplicate content. In addition, the model sometimes misinterprets extractive summarization as abstractive summarization. The most serious problem is that the model still pays too much attention to the context \textbf{at the beginning and end}. Our approach takes into account both intra-sentence and inter-sentence relationships, and effectively extracts key sentences distributed throughout the context and uses them as summaries. In addition, our model satisfies the input length constraints and saves computational resources.
\label{sec:llms}

\subsection{Human Evaluation}
We conduct human evaluation following the previous work \citep{luo2019reading}. We randomly sample 50 documents from the test sets of PubMed and Arxiv and ask three volunteers to evaluate the summaries extracted by HAESum, MTGNN, and LLM. For each document-summary pair, they are asked to rank them on three aspects: overall quality, coverage and non-redundancy. Notably the best one will be marked rank 1 and so on, and if both models extracted the same summaries they will both be ranked the same. We report the average results over the two datasets in Table \ref{tab8}

As seen through the table, our method achieves better results compared to other baselines. The human evaluation also further validates the effectiveness of our proposed method.
\label{sec:human}

\subsection{Distribution of Sentence Length and Number of Tokens in the Dataset}
In order to better demonstrate the validity of our choice of hyperparameters, we counted the distribution of sentence lengths in PubMed dataset as well as the distribution of the number of sentences. The experimental results are shown in Table \ref{tab9}

The obtained table shows that the hyperparameters we chose cover almost all the range of the distribution. This is further evidence that the choice of hyperparameters in the \textit{Hyperparameter sensitivity} section is adequate and effective.
\label{sec:distri}
\label{sec:appendix}


\end{document}